\documentclass[letterpaper, 10pt, conference]{ieeeconf}
\IEEEoverridecommandlockouts % allows \thanks
\usepackage{cite}
\usepackage{amsmath,amssymb}
\usepackage{amsthm}
\usepackage{algorithm}
\usepackage[noend]{algpseudocode}
\usepackage{float}
\usepackage{xcolor}
\usepackage{graphicx}
\usepackage{booktabs}
\usepackage[hyphens]{url}

\theoremstyle{plain}
\newtheorem{theorem}{Theorem}

\theoremstyle{definition}

\newtheorem{proposition}{Proposition}
\newtheorem{remark}{Remark}

\makeatletter
\newcommand{\fs@insetruled}{%
  \fs@ruled
  \def\@fs@pre{\kern4pt\hrule height.8pt depth0pt\kern2pt}%
}
\makeatother
\floatstyle{insetruled}
\restylefloat{algorithm}

\newcommand{\dur}[3]{D(#1,#2,#3)}
\definecolor{lsrpgray}{gray}{0.6}
\newcommand{\Old}{\color{lsrpgray}}   % lines inherited from LSRP*
\newcommand{\New}{\color{black}}      % new lines
\algrenewcommand\alglinenumber[1]{{\color{black}\footnotesize #1:}}
\newif\iffullversion
\fullversiontrue

\begin{document}

\title{\LARGE \bf
LSRP*-LA: Scalable and Anytime Multi-Agent Path Finding for Large Agents with Asynchronous Actions
}

\author{Dachuan Liu$^{1}$, Shizhe Zhao$^{1}$, and Zhongqiang Ren$^{1,\dagger}$%
\thanks{$^{1}$Dachuan Liu, Shizhe Zhao, and Zhongqiang Ren are with Shanghai Jiao Tong University, Shanghai, China {\tt\small dachuan@umich.edu, \{shizhe.zhao, zhongqiang.ren\}@sjtu.edu.cn}}%
\thanks{$^{\dagger}$Corresponding author.}%
}

\maketitle
\thispagestyle{empty}
\pagestyle{empty}

\begin{abstract}
Multi-Agent Path Finding (MAPF) seeks collision-free paths for multiple agents from their start locations to their goal locations while minimizing path costs. Most MAPF algorithms assume that each agent occupies a single vertex
and that the actions of all agents are synchronized. These assumptions may limit the use of MAPF planners in practice. Existing planners that get rid of both assumptions either handle only a few agents or cannot improve their solutions. Recently, a planner called LSRP* was proposed, which is scalable and anytime with asynchronous actions, but still limited by that each agent occupies a single vertex. This paper develops a new planner called LSRP*-LA for MAPF for large agents with asynchronous actions (LA-MAPF-AA), where an agent can occupy several cells. LSRP*-LA is scalable
and complete, and it keeps improving solution quality after the initial solution is obtained until the runtime budget depletes. We analyze the properties of LSRP*-LA and test it against several baselines with up to 800 agents on various maps. LSRP*-LA can handle twice as many agents as baseline methods and its solution cost is about half that of recent synchronous planners for large agents.
\end{abstract}

% Both label sets are kept, so \ref in the section files of either version resolves.
\section{Introduction}\label{lsrp-star-main:intro}
Multi-Agent Path Finding (MAPF) seeks collision-free paths for multiple agents from their start locations to their goal locations while minimizing the path costs. It arises in applications such as logistics and automated warehouses. MAPF is NP-hard to solve to optimality~\cite{yu2013structure},
and existing planners trade solution quality for scalability: optimal planners such as conflict-based search (CBS)~\cite{sharon2015cbs}, rule-based planners such as PIBT~\cite{okumura2022pibt}, which handle thousands of agents, and anytime planners, which keep improving a first solution while time
remains~\cite{li2021lns, okumura2023lacamstar}.

\begin{figure}[tb]
  \centering
  \includegraphics[width=\columnwidth]{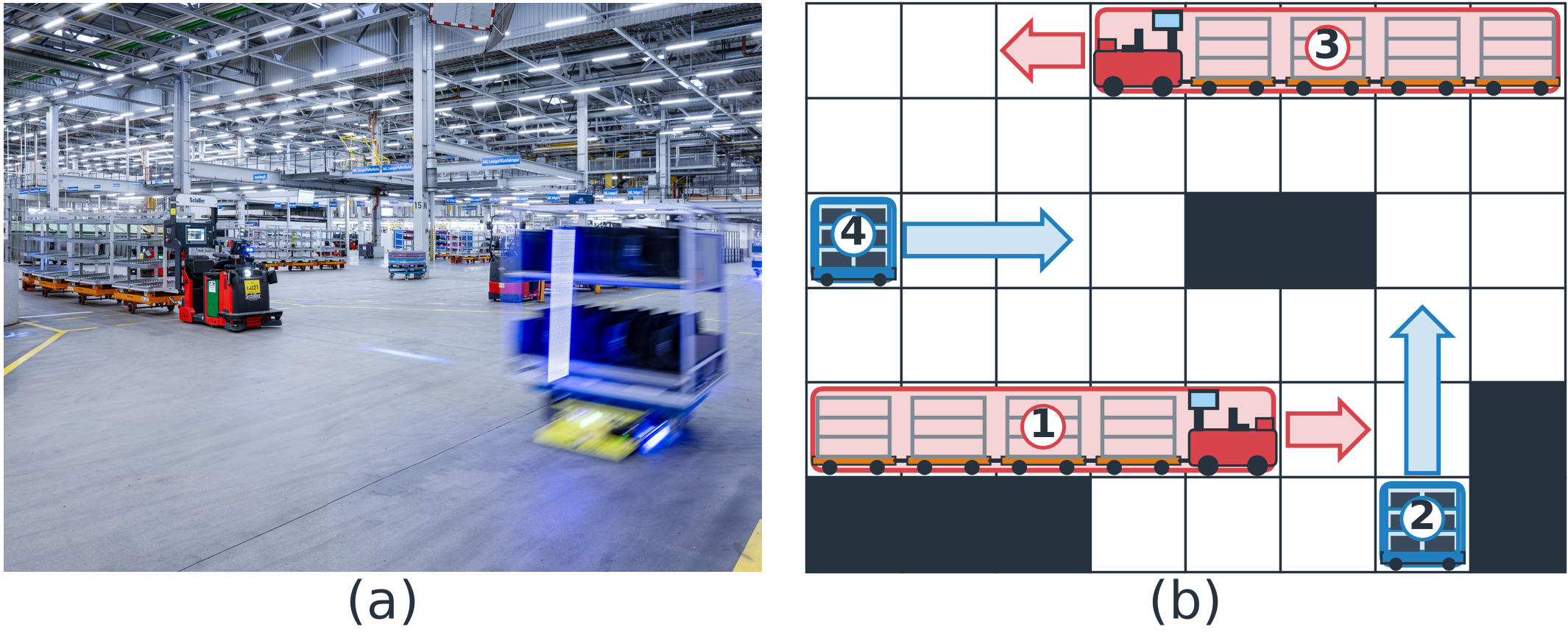}
  \caption{(a) Robots of different sizes and speeds at BMW Plant Regensburg.\protect\footnotemark[\value{footnote}] (b) Grid model: agents~1 and~3 (tugger trains) occupy five cells, agents~2 and~4 one cell; a longer arrow means a faster agent.}
  \label{fig:fig1}
\end{figure}

Most of these MAPF planners rely on two assumptions: (i) the point-agent assumption, where every agent occupies a single vertex of the graph, and (ii) the synchronous-action assumption, where every action of the agents, either wait or move, takes one time step, and all agents start and end their actions at the same integer time steps.
In practice, robot fleets may satisfy neither of the two assumptions. In the car plant of Fig.~\ref{fig:fig1}(a),
automated tugger trains and Smart Transport Robots (STRs) share one hall. A tugger train is several times longer
than an STR, so it occupies several cells and can block several STRs at the same time (Fig.~\ref{fig:fig1}(b)).
Tugger trains and STRs also move at different speeds, so they do not start and finish their moves at the same time.

Some existing work drops one of the two assumptions but keeps the other. In MAPF with asynchronous actions (MAPF-AA), every agent has its own action durations. Optimal planners for MAPF-AA solve instances with tens of agents~\cite{ren2021lss, wu2026cbsaa}. The rule-based planner LSRP~\cite{zhou2025lsrp} scales to hundreds of agents, and its extension LSRP*~\cite{liu2026lsrpstar} is in addition complete and anytime. In MAPF for large agents
(LA-MAPF)~\cite{li2019lamapf}, an agent occupies several cells, and even finding a feasible solution is
NP-hard~\cite{agafonov2025intractable}. Several variants of MAPF have been developed to address this problem, such as generalizing CBS~\cite{atzmon2020gmapf}, Multi-Dependency PIBT (MD-PIBT)~\cite{jiang2026mdpibt}, and
Layered LA-MAPF~\cite{yao2024layeredlamapf}. These planners assume synchronous actions.

Only a few planners drop both assumptions. CCBS~\cite{andreychuk2022ccbs} plans in continuous time for agents with a
geometric body that traverse edges of different lengths, and AOC-CBS~\cite{combrink2026aoccbs} handles
heterogeneous fleets and keeps improving its solution while time remains. These planners build on CBS and aim at
optimal solutions. Their runtime grows quickly with the number of conflicts, and large agents cause far more conflicts, leading to poor scalability. Prioritized planning with safe-interval path planning (PP-SIPP) handles robots of different sizes and speeds and scales to hundreds of robots~\cite{yakovlev2019diffdrive}, but it is incomplete and cannot improve its solution.

\stepcounter{footnote}
\footnotetext[\value{footnote}]{Image: BMW Group, \url{https://www.press.bmwgroup.com/global/article/detail/T0445860EN/}}

Different from these planners, this paper develops LSRP*-LA, a planner that drops both assumptions while being able to scale to a large number of agents with anytime solution refinement.
LSRP*-LA builds upon LSRP*~\cite{liu2026lsrpstar}, which does not consider large agents but can handle many point agents by employing a rule-based procedure to complete joint actions quickly, and a search over joint actions in the style of LaCAM~\cite{okumura2023lacam} to make the planner complete and anytime. 
However, extending LSRP* to large agents is challenging. First, a large agent may be blocked by several agents at once, and each of them may need several actions to get out of its way. Planning one action of a large agent thus couples many agents over several actions, and these
agents have to be planned together in their joint space, whose size grows exponentially with agents number. Second, due to asynchronous actions, the later actions of a push start at different times, and other agents are planned in between. Such an agent sees only the current footprints of the agents in the push, so it may move into the cells that one of their later actions will occupy. Synchronous planners for large agents (such as \cite{jiang2026mdpibt}) bypass this challenge by deciding the actions of all agents together at each time step.

To address these challenges, LSRP*-LA builds upon the search process of LSRP*, and checks collisions between the footprints of the agents. 
First, a pusher agent pushes all agents that block the pusher within an action, as in MD-PIBT~\cite{jiang2026mdpibt}, and each pushed agent plans a short sequence of actions, called an \emph{evacuation}, to leave the pusher's way. Second, a \emph{pre-execution check} tests each planned action for collisions right before it starts.
With these techniques, we show that LSRP*-LA is complete and guaranteed to eventually find an optimal solution given sufficient runtime.

In our tests, we compare LSRP*-LA against several baselines, including LSRP~\cite{zhou2025lsrp}, CBS-AA~\cite{wu2026cbsaa} extended to large agents, PP-SIPP~\cite{yakovlev2019diffdrive}, and the synchronous planners MD-PIBT~\cite{jiang2026mdpibt} and Layered LA-MAPF~\cite{yao2024layeredlamapf}, using various maps from a popular MAPF benchmark~\cite{stern2019mapf}. The results show that LSRP*-LA can handle up to 40 times as many agents as the baselines. Specifically, on den312d, our LSRP*-LA can solve instances with up to 800 agents, twice as many as LSRP alone, whereas CBS-AA and PP-SIPP solve no instance with more than 20 and 200 agents. The anytime search in our LSRP*-LA refines the cost of the initial solution by up to about 34\% within the runtime budget. LSRP*-LA often finds nearly optimal solutions, as its solution costs are at most 2\% higher than that of CBS-AA-LA, an optimal planner, on small scale problems. Planning with asynchronous actions is also important, as the solution costs of LSRP*-LA are about half of that of the synchronous planners (MD-PIBT~\cite{jiang2026mdpibt}, Layered LA-MAPF~\cite{yao2024layeredlamapf}) which ignores asynchronous actions.

\section{Related Work}
\textbf{Planning for MAPF-AA.}
Loosely Synchronized Search (LSS)~\cite{ren2021lss} and
CBS-AA~\cite{wu2026cbsaa} seek optimal solutions and handle tens of agents. Loosely Synchronized Rule-based
Planning (LSRP)~\cite{zhou2025lsrp} lets an agent push the agent that blocks it and scales to hundreds of agents,
but it cannot improve its solution. LSRP*~\cite{liu2026lsrpstar} makes
LSRP complete and anytime with a search inspired by LaCAM*~\cite{okumura2023lacam,okumura2023lacamstar}. LaCAM*
keeps only the cheapest path to each joint configuration, which does not suffice with asynchronous actions. LSRP* therefore searches over joint actions together with their start and end times, and
compares two paths by both the cost and the arrival times of the agents. These planners assume that each agent occupies a single vertex.

\textbf{Planning for large agents.}
CBS has been extended to large agents~\cite{li2019lamapf, atzmon2020gmapf}. Multi-Dependency PIBT
(MD-PIBT)~\cite{jiang2026mdpibt} lets one agent push several agents, and Layered
LA-MAPF~\cite{yao2024layeredlamapf} splits an instance into smaller ones. These planners assume synchronous
actions. HetPIBT~\cite{chakravarty2026hetpibt} extends PIBT to robots of different sizes and speeds, but it is incomplete and cannot improve its solution.

\section{Problem Definition}\label{lsrp-star-main:problem-def}
Let $I =\{1,2,\dots,N\}$ denote a set of $N$ agents. All agents move in a workspace modeled by a finite graph
$G =(V,E)$, where the vertices $V$ represent all possible locations of the agents and the edges
$E\subseteq V\times V$ denote the pairs of adjacent vertices between which agents can move.

In MAPF for large agents, the workspace is divided into unit squares, called
\emph{cells}. An agent occupies a set of cells. Let $Q$ denote the finite set of cells and $O\subseteq Q$ the set of \emph{obstacle cells}; the other
cells are \emph{free cells}. Every vertex $v\in V$ is the center of a free cell. When agent $i\in I$ is at vertex
$v$, it occupies the set of cells $F^i(v)\subseteq Q$, called its \emph{footprint}. When agent $i$ moves from $u$ to $v$, it occupies the set of
cells $F^i_{sw}(u,v)\subseteq Q$, called its \emph{swept area}. It satisfied $F^i(u)\cup F^i(v)\subseteq F^i_{sw}(u,v)$ and $F^i_{sw}(v,v)=F^i(v)$ when $i$ stays at $v$.

Two sets of
cells $X$ and $Y$ \emph{collide} if they share a cell, $X\cap Y\neq\emptyset$. Agent $i$ moves on the subgraph
$G^i=(V^i,E^i)$ of the positions and moves on free cells:
\begin{align*}
  V^i&=\{v\in V\mid F^i(v)\subseteq Q\setminus O\},\\
  E^i&=\{(u,v)\in E\mid F^i_{sw}(u,v)\subseteq Q\setminus O\}.
\end{align*}
Locations and moves are discrete, and time is continuous. For each agent $i$, let $\dur{i}{u}{v}\in\mathbb{R}^+$ be the duration of a move along $(u,v)\in E^i$.
Different agents may have different durations for the same edge.

An \emph{action} is a tuple $a^i=(p,v,t_p,t_v)$.We use the \emph{parenthesis operator} to access a component of a tuple: given a tuple $y$ and one of its components $x$, let $x(y)$ denote that component of $y$; e.g., $t_v(a^i)$ is the arrival time of $a^i$. Agent $i$ departs from $p$ at time $t_p$ and arrives at $v$ at time $t_v$
along $(p,v)\in E^i$. It is a \emph{wait} if $p=v$, of any positive duration $t_v-t_p$, and a \emph{move} if
$p\neq v$, with $t_v-t_p=\dur{i}{p}{v}$. In both cases, agent $i$ occupies the swept area $F^i_{sw}(p,v)$ during the interval $[t_p,t_v)$; for a wait, this is its footprint $F^i(v)$. Two action $a^i$ and $a^j$ of different agents \emph{conflict} if their swept areas collide and their time intervals overlap:
\begin{align}
  &F^i_{sw}\big(p(a^i),v(a^i)\big)\cap F^j_{sw}\big(p(a^j),v(a^j)\big)\neq\emptyset \nonumber\\
  &\wedge\;[t_p(a^i),t_v(a^i))\cap[t_p(a^j),t_v(a^j))\neq\emptyset .
  \label{eq:conflict}
\end{align}

Let $v^i_s,v^i_g\in V^i$ denote the start and goal vertices of agent $i$. The start footprints and goal footprints of different agents do not
collide: $F^i(v^i_s)\cap F^j(v^j_s)=\emptyset$ and
$F^i(v^i_g)\cap F^j(v^j_g)=\emptyset$ for all $i\neq j$. All agents share a global clock and start at time $0$. A
\emph{path} of agent $i$ is a sequence of consecutive actions $\pi^i=\langle a^i_0,\dots,a^i_K\rangle$, i.e.,
$v(a^i_k)=p(a^i_{k+1})$ and $t_v(a^i_k)=t_p(a^i_{k+1})$, from $v^i_s$ at time $0$ to $v^i_g$. Afterwards, $i$
stays at its goal, which we treat as a final wait $(v^i_g,v^i_g,t_v(a^i_K),\infty)$. For a prefix of $\pi^i$ ending
at action $a^i_k$, its cost is
\begin{equation}
g(a^i_k) =
\begin{cases}
g(a^i_{k-1}), & k>0 \text{ and } p(a^i_k)=v(a^i_k)=v^i_g,\\
t_v(a^i_k), & \text{otherwise},
\end{cases}
\label{eq:cost}
\end{equation}
so it stays frozen while $i$ waits at its goal. The cost of the path is $g(\pi^i)=g(a^i_K)$. A \emph{joint path}
$\pi=(\pi^1,\dots,\pi^N)$ is a \emph{solution} if no two actions of different agents conflict, final waits
included. We denote $g(\pi)=\sum_{i=1}^N g(\pi^i)$ the \emph{sum of costs} (SOC).

The goal of Multi-Agent Path Finding for Large Agents with Asynchronous Actions (LA-MAPF-AA) is to find a
conflict-free joint path $\pi$ connecting $v^i_s,v^i_g$ for all agents $i\in I$, such that $g(\pi)$ reaches the
minimum.

\begin{remark}\label{rem:practice}
In practice, the footprint of a robot or an obstacle may cover only part of a cell. The whole cell then counts as occupied: it is an obstacle cell, or it belongs to the footprint or the swept area of the robot. Hence no other agent may enter any part of this cell while it is occupied.
\end{remark}

\begin{remark}\label{rem:special}
If every footprint is one cell and $F^i_{sw}(u,v)=F^i(u)\cup F^i(v)$ for every move, a conflict is a vertex-level conflict, and LA-MAPF-AA becomes MAPF-AA~\cite{ren2021lss}. If every move and every wait lasts one time unit, all
actions are synchronous, and LA-MAPF-AA becomes LA-MAPF~\cite{li2019lamapf} in the variant where a move occupies its
swept area during the whole time step.
\end{remark}

% \section{Background}\label{lsrp-star-background}
% \input{latex/background}

\section{Method}\label{lsrp-star-main:method}
\begin{figure}[tb]
  \centering
  \includegraphics[width=\columnwidth]{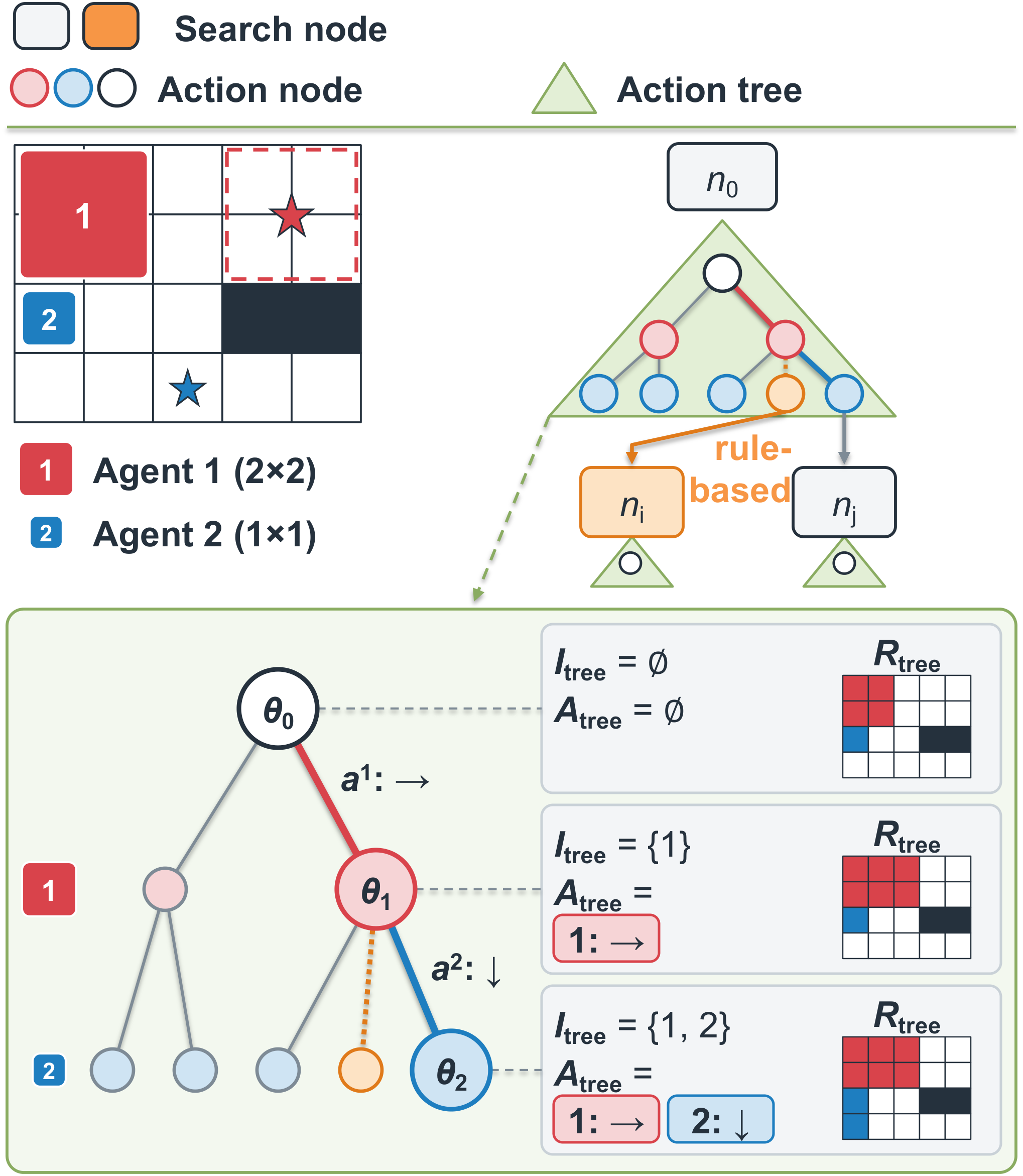}
  \caption{Search tree and action tree of LSRP*-LA with a $2{\times}2$ agent~1 and a $1{\times}1$ agent~2 (stars:
  goals). Top: each time a search node is selected, one successor is generated from one action node of its action tree; $n_j$ comes from the leaf $\theta_2$, and $n_i$ (orange) from $\theta_1$, whose agent~2 is completed by the rule-based procedure.
}
  \label{fig:lsrpstar}
\end{figure}

\subsection{Notations and Concepts}\label{sec:bg:joint} 

\subsubsection{\textbf{Joint Action}} As agents start and end their actions at different times, a \emph{joint action} $a=(a^i)_{i\in I}$ contains one action per agent, all in progress at a common time. Its  \emph{current event} $\tau(a)=\min_{i}t_v(a^i)$ is the earliest arrival; the agents $I_{\text{curr}}(a)=\{i\mid t_v(a^i)=\tau(a)\}$ are \emph{idle} at $\tau(a)$ and the others are \emph{busy}. Let $v^i=v(a^i)$ denote the vertex at which agent $i$ arrives. A \emph{successor} $a_{\text{next}}$ of $a$ gives every idle agent $i$ a new action that starts at $\tau(a)$, a move along $(v^i,u)\in E^i$ or a wait at $v^i$,
and keeps the actions of the busy agents; its waits end at its \emph{next event} $\tau_{\text{next}}(a)$, the earliest arrival of a move in $a_{\text{next}}$. A dummy joint action $a_0$ with $a^i_0=(v^i_s,v^i_s,0,0)$ makes all agents idle at time~0, and a joint path is a sequence $\langle a_0,\dots,a_K\rangle$ in which each joint action is a successor of the previous one.

\subsubsection{\textbf{Action tree}} A successor of $a$ is \emph{valid} if no two of its actions conflict. The number of valid successors can be exponential in the number of idle agents $m=|I_{\text{curr}}(a)|$. The \emph{action tree} $AT(a)$ enumerates them all and the leaves are exactly the valid successors of $a$ (Fig.~\ref{fig:lsrpstar}, bottom). Let $i_1,\dots,i_m$ be the idle agents. A node of $AT(a)$ is called an \emph{action node}. An action node $\theta$ of depth $d$
assigns the actions $A_{\text{tree}}(\theta)$ to the agents $I_{\text{tree}}(\theta)=\{i_1,\dots,i_d\}$. The \emph{root action node} $\theta_0$ assigns nothing. $R_{\text{tree}}(\theta)$ denotes the cells that the agents occupy by $\theta$: the swept areas of the assigned actions and of the busy agents' actions, and the footprints of the unassigned idle agents. The children $\mathrm{Ch}^{LA}(\theta)$ of $\theta$ assign one more action to agent $i=i_{d+1}$: a wait, or a move to a vertex $u$ with $\big(F^i_{sw}(v^i,u)\setminus F^i(v^i)\big)\cap R_{\text{tree}}(\theta)=\emptyset$. In Fig.~\ref{fig:lsrpstar}, agent~1 can wait or move right at $\theta_0$; moving down would sweep the cell of agent~2. Moving right gives $\theta_1$. Under $\theta_1$, agent~2 can wait, move right or move down, and moving down gives the leaf $\theta_2$.

\subsubsection{\textbf{Search tree}} The leaves of the action tree $AT(a)$ are the valid successors of $a$. Generating them recursively from the initial joint action $a_0$ forms the \emph{search tree}. A \emph{search node} $n=(a,n_{\text{parent}},\mathcal{T},\lambda)$ stores a joint action, its parent, the \emph{frontier} $\mathcal{T}$ of its action tree, i.e., the action nodes not yet expanded, and a \emph{rule-based state} $\lambda$ (Sec.~\eqref{method:notations:evacuation}). A successor of $n$ stores a valid successor of $a(n)$. Its cost is $g(n)=\sum_{i}g(a^i(n))$
(Eq.~\eqref{eq:cost}), and $f(n)=g(n)+\sum_{i}h^i(v^i)$ is a lower bound on the SOC of every solution through $n$, where $h^i(u)$ is the duration of a fastest path from $u$ to $v^i_g$ in $G^i$. A \emph{goal node} $n_g$ has $v(a^i(n_g))=v^i_g$ for all $i$, and the joint path from $a_0$ to $a(n_g)$ is a solution.

\subsubsection{\textbf{Push Tree and Ban Chain}}
\begin{figure}[tb]
  \centering
  \includegraphics[width=\columnwidth]{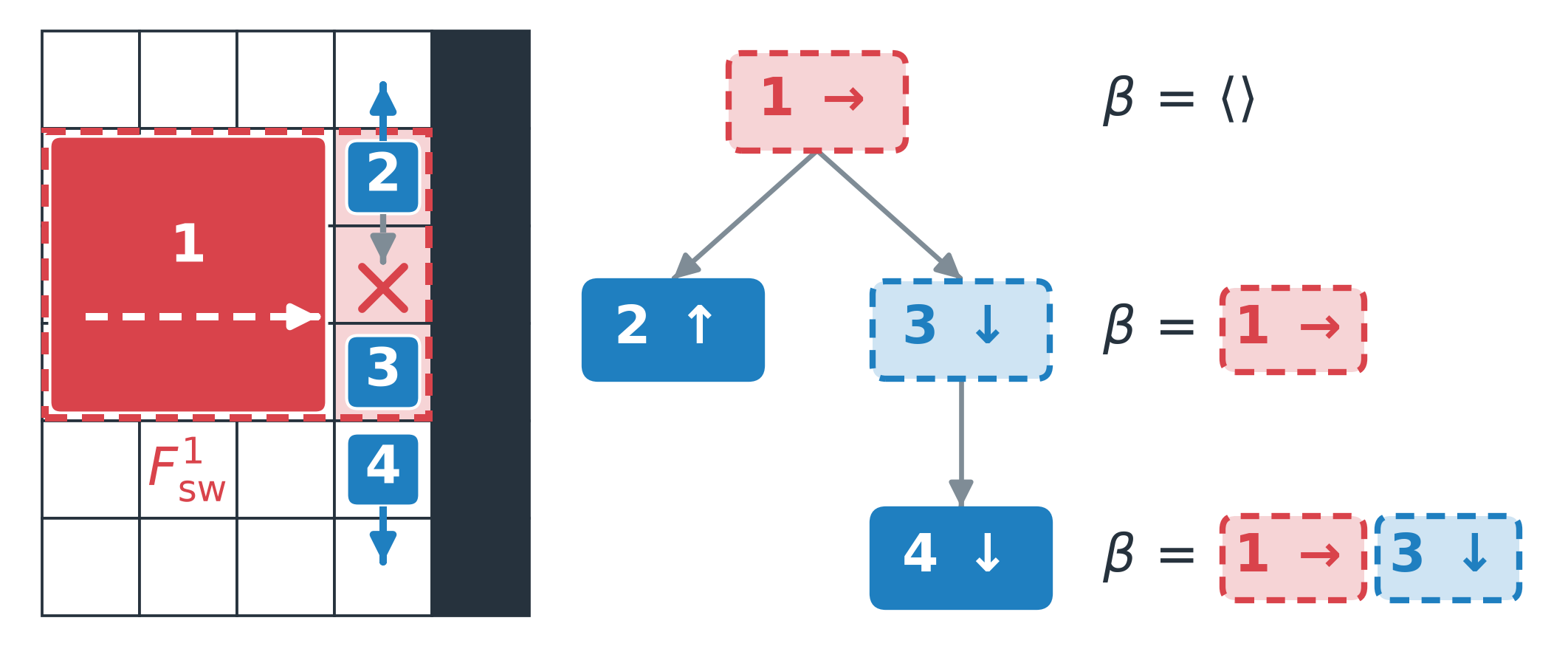}
  \caption{Push tree and ban chain. The $3{\times}3$ agent~1 tries to move right, and its swept area $F^1_{sw}$ (red
  dashed) covers agents~2 and~3. $\times$: move forbidden by the ban chain. Right: the push tree; solid nodes move
  now, and dashed nodes wait for their release times.}
  \label{fig:push_tree}
\end{figure}
Each agent $i$ has a \emph{priority} $\epsilon^i$, initially in $(0,1)$. At every successor, $\epsilon^i$ grows by one, unless $i$ is at its goal, where it is reset to its initial value. An agent is \emph{planned} once its action in the successor is assigned. The rule-based procedure plans the unplanned idle agents in descending priority.

In LSRP*, an agent \emph{pushes} the unplanned agent at its next vertex and moves after that agent has left. A large agent may have to push several agents. Let agent $i$ try to move from $v^i$ to a neighbor $v$, and let $B$ be the set
of unplanned agents whose footprints collide with $F^i_{sw}(v^i,v)$. If $B$ is empty, $i$ moves now. Otherwise, $i$ waits at $v^i$ and pushes every agent in $B$, which may push other agents in turn. These pushes form the \emph{push tree} (Fig. \ref{fig:push_tree}): its root is the first pusher, and the children of an agent are the agents that it pushes. A waiting pusher is planned, so no agent pushes it back. Agent $i$ moves to $v$ at its \emph{release time} $t^{\text{rel}}$,
when all agents in $B$ have left $F^i_{sw}(v^i,v)$; this move is a \emph{deferred move}. If an agent in $B$ finds no way out, the push fails, and $i$ tries another vertex.

A pushed agent must not stop in the swept area of a pusher above it in the push tree, since it would block that pusher again. The \emph{ban chain} $\beta=\langle(k_1,u_1),\dots,(k_r,u_r)\rangle$ of a pushed agent lists its
ancestors $k_q$ in the push tree, each with the destination $u_q$ of its deferred move. Their swept areas form $S(\beta)=\bigcup_{q=1}^{r}F^{k_q}_{sw}(v^{k_q},u_q)$. A pushed agent may only stop at a vertex whose footprint does not collide with $S(\beta)$. It may cross $S(\beta)$ on the way, because its pushers start moving only after it has arrived. The ban chain of an agent that nobody pushes is empty, $\langle\rangle$. The ban chain of an agent pushed by $i$ is the ban chain of $i$ with $(i,v)$ appended.

In Fig.~\ref{fig:push_tree}, agent~1 pushes agents~2 and~3, which get $\beta=\langle(1,u_1)\rangle$ and $S(\beta)=F^1_{sw}$. Agent~2 cannot move down~($\times$) and moves up. Agent~3 cannot move up and pushes agent~4 down; agent~4 has $\beta=\langle(1,u_1),(3,u_3)\rangle$. Agent~3 moves after agent~4 has left, and agent~1 after agents~2 and~3 have left.

\subsubsection{\textbf{Evacuation and Action Queue}}\label{method:notations:evacuation}
\begin{figure}[tb]
  \centering
  \includegraphics[width=\columnwidth]{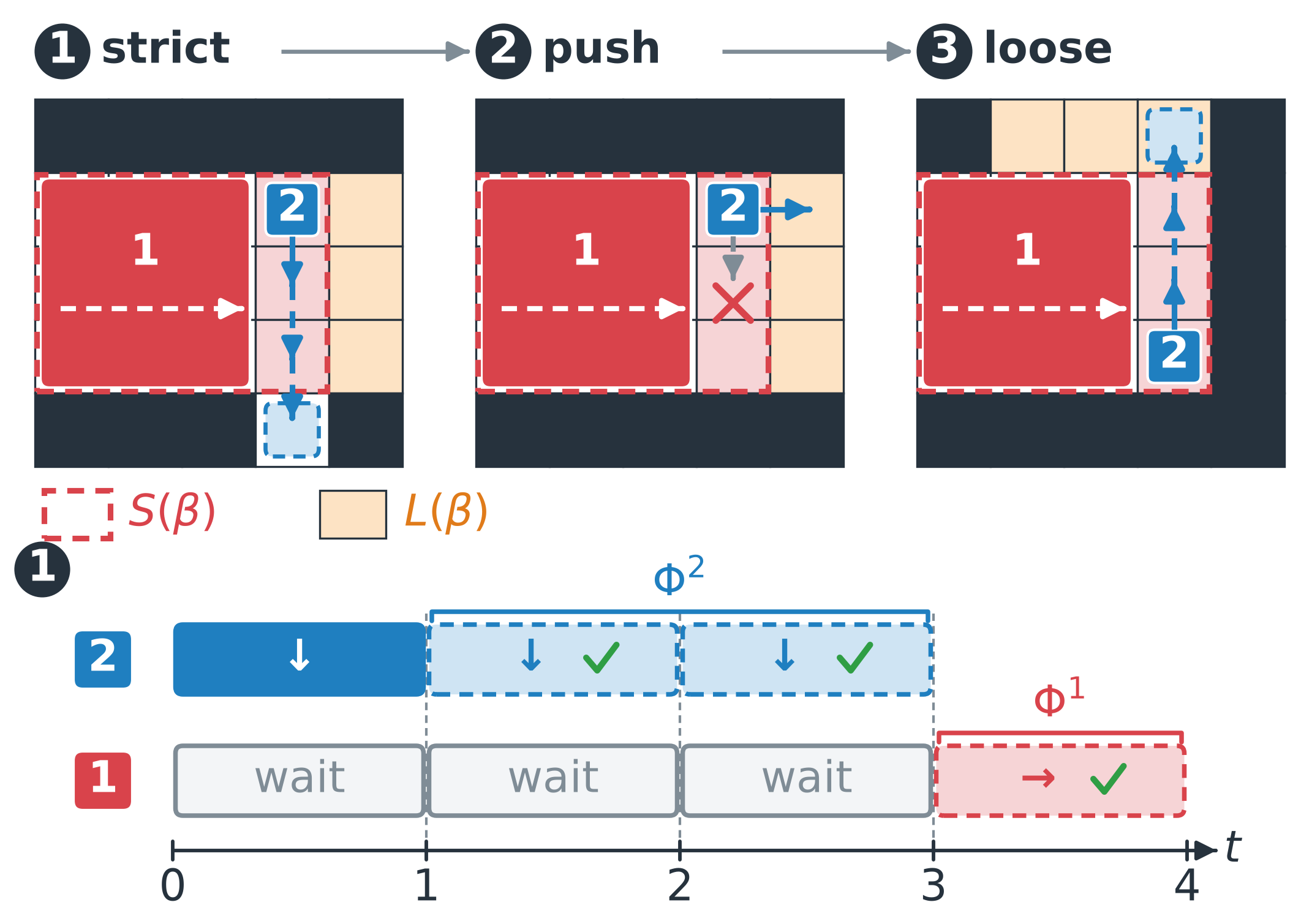}
  \caption{Ways out for agent~2, pushed by the $3{\times}3$ agent~1, tried in the order 1--3. (1)~Strict
  evacuation. (2)~Single move; $\times$: it may not end in $S(\beta)$. (3)~Loose evacuation. Dashed moves are
  stored in $\Phi$. Bottom: time line of~(1); \checkmark: pre-execution check.}
  \label{fig:evacuation_queue}
\end{figure}
A pushed agent must stop outside $S(\beta)$. With a large pusher, this may take several moves. A single move may end outside $S(\beta)$ but in the way of pusher's later moves, so the pusher pushes the agent again. We therefore let a pushed agent take a short path of several moves, called an \emph{evacuation}.

Let $j$ be a pushed agent with ban chain $\beta$, and let $(i,u)$ be the last entry of $\beta$, i.e., agent $i$ pushed $j$ and waits to move to $u$. Let $w_0=u$, and let $w_{q+1}$ be the neighbor of $w_q$ in $G^i$ with the smallest $h^i$. The \emph{lane} $L(\beta)=\bigcup_{q=0}^{\ell}F^i(w_q)$ is the set of cells that $i$ occupies in its next $\ell$ moves on a shortest path to its goal, where $\ell$ is a parameter. It is only a guess, as it ignores the other agents.

An evacuation of $j$ is a path of several moves in $G^j$ on which $j$ pushes nobody: no move on it may collide with the swept area of a planned agent or the footprint of another agent. Agent $j$ tries three ways out in this order
(Fig.~\ref{fig:evacuation_queue}), and the push fails if all three fail. (1)~\emph{Strict evacuation}: an evacuation
of at most $d_s$ moves that ends outside both $S(\beta)$ and $L(\beta)$, so that $i$ will not push $j$ again soon.
(2)~A single move that ends outside $S(\beta)$, as in the push tree; it keeps $j$ near its shortest path but may leave it on the lane. (3)~\emph{Loose evacuation}: an evacuation of at most $d_l$
moves that ends outside $S(\beta)$. It is the last resort when no single move leaves $S(\beta)$, and it at least lets $i$ move. Here $d_s$ and $d_l$ are parameters.

The first move of an evacuation starts now. Its later moves and the deferred move of a pusher start at later events,
so each agent $i$ keeps them in an \emph{action queue} $\Phi^i$, in the order of execution. At a later event, an idle agent $i$ executes the first action of $\Phi^i$ if it is due, after a \emph{pre-execution check}: the swept area of the action must collide with no footprint or swept area of another agent. This check is new for large agents: while $i$ waits, other agents may enter the unoccupied cells of its future swept area. If the check fails, $i$ drops $\Phi^i$ and is planned again.

The priorities and the action queues form the \emph{rule-based state} $\lambda=(\epsilon,\Phi)$. Each search node $n$ stores $\lambda(n)$ and passes it to its successors.

\begin{figure}[tb]
  \centering
  \includegraphics[width=\columnwidth]{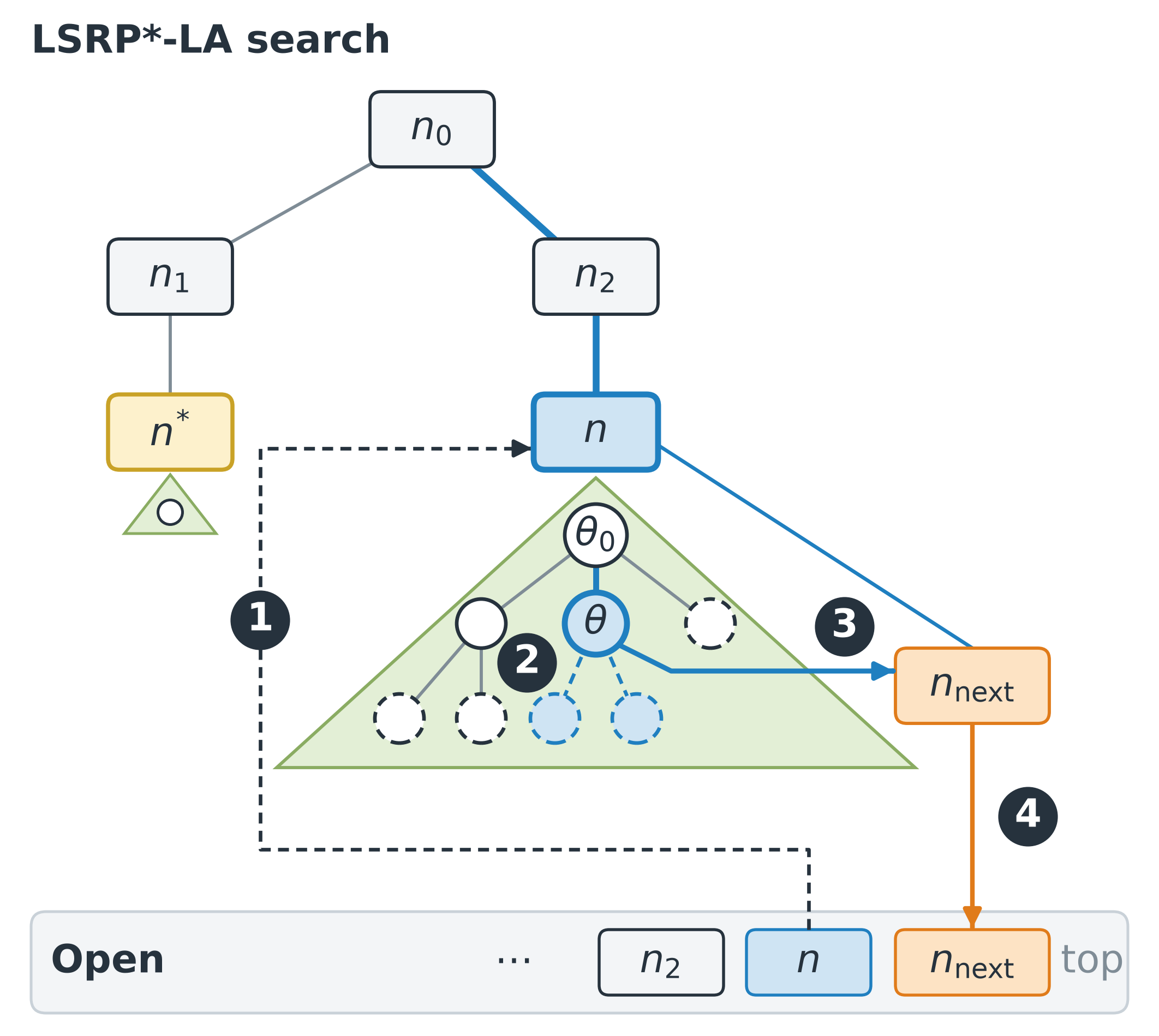}
  \caption{One iteration of the LSRP*-LA search. (1)~Pop $n$ from $Open$. (2)~Expand the first action node
  $\theta$ of its frontier. (3)~Generate $n_{\text{next}}$ from $\theta$. (4)~Push $n$ and $n_{\text{next}}$.
  Gold: the incumbent $n^*$.}
  \label{fig:search}
\end{figure}

\subsection{Algorithm Overview}\label{sec:overview}
 This work develops LSRP*-LA that can quickly solve LA-MAPF-AA instances with many agents by finding a sub-optimal solution. LSRP*-LA has two parts. The \emph{search algorithm} (Sec.~\ref{sec:search}) builds the search tree and keeps improving the best solution found so far. It makes LSRP*-LA complete and anytime. The \emph{rule-based procedure}
(Sec.~\ref{sec:method:rule}) quickly plans the idle agents that the search leaves unassigned. It makes LSRP*-LA find
a first solution fast. Large agents change the search only through $\mathrm{Ch}^{LA}$. All other changes lie in the
rule-based procedure.

\subsection{Search Algorithm}\label{sec:search}
LSRP*-LA searches as LSRP* does (Fig.~\ref{fig:search}). It keeps the search nodes in a stack $Open$, which
initially holds the root $n_0$ with $a_0$ and $\mathcal{T}(n_0)=\{\theta_0\}$. It repeats four steps.
(1)~It pops the top search node $n$ of $Open$.
(2)~It removes the first action node $\theta$ from the frontier $\mathcal{T}(n)$ and appends the children
$\mathrm{Ch}^{LA}(\theta)$.
(3)~It generates one successor $n_{\text{next}}$ of $n$: the busy agents keep their actions, the agents in
$I_{\text{tree}}(\theta)$ take the actions in $A_{\text{tree}}(\theta)$, and the rule-based procedure plans the other
idle agents.
(4)~It pushes $n$ back if $\mathcal{T}(n)\neq\emptyset$, and then pushes $n_{\text{next}}$ with
$\mathcal{T}(n_{\text{next}})=\{\theta_0\}$.
Since $n_{\text{next}}$ is on top of $Open$, the search goes deep and quickly reaches a goal node. The
\emph{incumbent} $n^*$ is the cheapest goal node found so far. Once it exists, a popped search node $n$ with
$f(n)\ge g(n^*)$ is pruned. The search returns $n^*$ when $Open$ is empty or time runs out.

\begin{algorithm}[tb]
\small
\caption{\textsc{LSRPStep-LA}.}\label{alg:lsrpstep-la}
\begin{algorithmic}[1]
\Old
\Procedure{LSRPStep-LA}{$n, \partial a_{\text{next}}, isRoot$} \label{alg:lsrpstep-la:proc}
    \State $(\epsilon, \Phi) \gets \lambda(n)$; update the priorities $\epsilon$ \label{alg:lsrpstep-la:init}
    \If{\textbf{not} $isRoot$} $\Phi \gets \emptyset$ \Comment{only the root reads $\Phi$} \EndIf \label{alg:lsrpstep-la:nonroot}
    \For{$i \in I_{\text{curr}}(a(n))$ with $\Phi^i \neq \emptyset$} \label{alg:lsrpstep-la:queue_loop}
        \New
        \State $\varphi \gets \Phi^i.\text{top}()$ \label{alg:lsrpstep-la:top}
        \Old
        \If{$t_p(\varphi) > \tau$} $i$ waits at $v^i$ \Comment{not due yet} \label{alg:lsrpstep-la:wait}
        \New
        \ElsIf{$t_p(\varphi) = \tau \land F^i_{sw}(p(\varphi),v(\varphi)) \cap \mathcal{C}^{-i} = \emptyset$} \label{alg:lsrpstep-la:check}
            \State $\partial a^i_{\text{next}} \gets \varphi$; $\Phi^i.\text{pop}()$ \Comment{pre-execution check} \label{alg:lsrpstep-la:execute}
        \Else\ $\Phi^i.\text{clear}()$ \Comment{start time passed or check failed} \label{alg:lsrpstep-la:drop}
        \EndIf
        \Old
    \EndFor
    \For{$i \in I_{\text{curr}}(a(n))$ in descending $\epsilon^i$} \label{alg:lsrpstep-la:agent_loop}
        \If{$i$ is unplanned} \textsc{AsyPush-LA}$(i, \langle\rangle)$ \EndIf \label{alg:lsrpstep-la:asypush}
    \EndFor
    \State \Return $\big((\epsilon, \Phi), \partial a_{\text{next}}\big)$ \label{alg:lsrpstep-la:return}
\EndProcedure
\New
\end{algorithmic}
\end{algorithm}

\subsection{Rule-Based Procedure}\label{sec:method:rule}
\begin{algorithm}[tb]
\small
\caption{\textsc{AsyPush-LA}.}\label{alg:asypush-la}
\begin{algorithmic}[1]
\New
\Procedure{AsyPush-LA}{$i, \beta$} \Comment{$\beta$: ban chain} \label{alg:asypush-la:proc}
    \If{$\beta \neq \langle\rangle$} \Comment{strict evacuation} \label{alg:asypush-la:check_strict}
        \State $P \gets \textsc{EscPath}(i, \beta, L(\beta), d_s)$ \label{alg:asypush-la:strict}
        \If{$P \neq \emptyset$} \Return \textsc{TakePath}$(i, P)$ \EndIf \label{alg:asypush-la:take_strict}
    \EndIf
    \Old
    \For{$v \in \textsc{Order}(i, \partial a_{\text{next}})$} \label{alg:asypush-la:cand_loop}
        \New
        \If{$F^i_{sw}(v^i, v) \cap \mathcal{C}_{\text{pl}} \neq \emptyset$} \textbf{continue} \EndIf \label{alg:asypush-la:check_planned}
        \If{$F^i(v) \cap S(\beta) \neq \emptyset$} \textbf{continue} \Comment{ban chain} \EndIf \label{alg:asypush-la:check_ban}
        \State $B \gets \{\text{unplanned } j \neq i \mid F^j(v^j) \cap F^i_{sw}(v^i, v) \neq \emptyset\}$ \label{alg:asypush-la:blockers}
        \Old
        \If{$B = \emptyset$} $i$ moves to $v$ now; \Return $t_v(\partial a^i_{\text{next}})$ \EndIf \label{alg:asypush-la:move}
        \State $i$ waits at $v^i$ \label{alg:asypush-la:wait}
        \New
        \State $\beta' \gets \beta$ extended by $(i, v)$ \Comment{$i$ becomes a pusher} \label{alg:asypush-la:ban}
        \State $t^{\text{rel}} \gets \max_{j \in B} \textsc{AsyPush-LA}(j, \beta')$ \Comment{$\bot$ if one fails} \label{alg:asypush-la:push}
        \Old
        \If{$t^{\text{rel}} = \bot$} $\partial a^i_{\text{next}} \gets \emptyset$; \textbf{continue} \EndIf \label{alg:asypush-la:fail}
        \State $\Phi^i \gets \langle (v^i, v, t^{\text{rel}}, \bot) \rangle$ \Comment{deferred move} \label{alg:asypush-la:defer}
        \State \Return $t^{\text{rel}} + \dur{i}{v^i}{v}$ \label{alg:asypush-la:return_defer}
    \EndFor
    \New
    \If{$\beta \neq \langle\rangle$} \Comment{loose evacuation} \label{alg:asypush-la:check_loose}
        \State $P \gets \textsc{EscPath}(i, \beta, \emptyset, d_l)$ \label{alg:asypush-la:loose}
        \If{$P \neq \emptyset$} \Return \textsc{TakePath}$(i, P)$ \EndIf \label{alg:asypush-la:take_loose}
    \EndIf
    \Old
    \State $i$ waits at $v^i$; \Return $\bot$ \label{alg:asypush-la:return_fail}
\EndProcedure
\New
\end{algorithmic}
\end{algorithm}

Algs.~\ref{alg:lsrpstep-la}--\ref{alg:evacuation} give the rule-based procedure. In Algs.~\ref{alg:lsrpstep-la} and~\ref{alg:asypush-la}, gray lines are as in \textsc{LSRPStep} and \textsc{AsyPush}
of LSRP*~\cite{liu2026lsrpstar}, and black lines are new for LSRP*-LA. 
Let $\tau=\tau(a(n))$ denote the current event. The \emph{partial joint action} $\partial a_{\text{next}}$ holds the actions assigned so far, initialized by actions of busy agents and $A_{\text{tree}}(\theta)$. Agent $j$ is planned if
$\partial a^j_{\text{next}}$ is set. All procedures share $\partial a_{\text{next}}$ and the action queues $\Phi$ and
change them in place. Let $\mathcal{C}_{\text{pl}}=\bigcup_{j\text{ planned}}F^j_{sw}\big(p(\partial a^j_{\text{next}}),v(\partial a^j_{\text{next}})\big)$
denote the swept area occupied by the planned agents, and
$\mathcal{C}^{-i}=\mathcal{C}_{\text{pl}}\cup\bigcup_{j\neq i\text{ unplanned}}F^j(v^j)$ the cells occupied by all
agents other than $i$. The symbol $\bot$ marks a failure, or an arrival time that is not set yet.

\textbf{Collective step.} \textsc{LSRPStep-LA} (Alg.~\ref{alg:lsrpstep-la}) updates the priorities, runs the action
queues, and then plans the unplanned idle agents in descending priority. The flag $isRoot$ is true if
$\theta=\theta_0$. As in LSRP*, the action queues are used only in this case. Otherwise, $\theta$ has already
assigned some actions, and the queues are dropped (line~\ref{alg:lsrpstep-la:nonroot}). Let $\varphi$ be the first
action of $\Phi^i$. If $\varphi$ starts after $\tau$, $i$ waits (line~\ref{alg:lsrpstep-la:wait}). If $\varphi$
starts at $\tau$ and passes the pre-execution check, i.e., $F^i_{sw}(p(\varphi),v(\varphi))$ does not collide with
$\mathcal{C}^{-i}$, $i$ executes $\varphi$ (lines~\ref{alg:lsrpstep-la:check}--\ref{alg:lsrpstep-la:execute}). In
all other cases, $\Phi^i$ is cleared (line~\ref{alg:lsrpstep-la:drop}), and $i$ is planned by \textsc{AsyPush-LA}
(line~\ref{alg:lsrpstep-la:asypush}).

\textbf{Single-agent step.} \textsc{AsyPush-LA}$(i,\beta)$ (Alg.~\ref{alg:asypush-la}) plans agent $i$ with ban
chain $\beta$. A pushed agent tries the three ways out of Fig.~\ref{fig:evacuation_queue} in order: strict
evacuation (lines~\ref{alg:asypush-la:check_strict}--\ref{alg:asypush-la:take_strict}), a single move
(lines~\ref{alg:asypush-la:cand_loop}--\ref{alg:asypush-la:return_defer}), and loose evacuation
(lines~\ref{alg:asypush-la:check_loose}--\ref{alg:asypush-la:take_loose}). An agent that nobody pushes
($\beta=\langle\rangle$) tries only single moves. \textsc{AsyPush-LA} returns the time at which $i$ arrives outside
$S(\beta)$, the swept areas of its pushers, or $\bot$ if all ways fail (line~\ref{alg:asypush-la:return_fail}). For a single move, \textsc{Order} sorts $v^i$ and its neighbors in $G^i$ by how fast $i$ can reach its goal through
them, as in LSRP*. A vertex $v$ is skipped if $F^i_{sw}(v^i,v)$ collides with $\mathcal{C}_{\text{pl}}$
(line~\ref{alg:asypush-la:check_planned}), or if $F^i(v)$ collides with $S(\beta)$
(line~\ref{alg:asypush-la:check_ban}). Let $B$ be the set of unplanned agents in the way of the move, i.e., those
whose footprints collide with $F^i_{sw}(v^i,v)$ (line~\ref{alg:asypush-la:blockers}). If $B$ is empty, $i$ moves to
$v$ now (line~\ref{alg:asypush-la:move}). Otherwise, $i$ waits and pushes the agents in $B$ with the ban chain
$\beta'$, i.e., $\beta$ with $(i,v)$ appended (lines~\ref{alg:asypush-la:wait}--\ref{alg:asypush-la:push}). Its move
to $v$ is deferred to the release time $t^{\text{rel}}$, the latest time that these pushes finish, and goes into $\Phi^i$ (lines~\ref{alg:asypush-la:defer}--\ref{alg:asypush-la:return_defer}). If a push fails, $i$ tries the next
vertex, and the agents already pushed keep their new actions (line~\ref{alg:asypush-la:fail}).

\begin{algorithm}[tb]
\small
\caption{\textsc{EscPath} and \textsc{TakePath}.}\label{alg:evacuation}
\begin{algorithmic}[1]
\Procedure{EscPath}{$i, \beta, L, d_{\max}$} \label{alg:evacuation:esc_proc}
    \State $E' \gets \{(u, w) \in E^i \mid F^i_{sw}(u, w) \cap \mathcal{C}^{-i} = \emptyset\}$ \label{alg:evacuation:edges}
    \State $V_{\text{reach}} \gets$ vertices within $d_{\max}$ moves of $v^i$ over $E'$
    \label{alg:evacuation:bfs}
    \State $U \gets \{u \in V_{\text{reach}} \mid F^i(u) \cap (S(\beta) \cup L) = \emptyset\}$ \label{alg:evacuation:goal}
    \If{$U = \emptyset$} \Return $\emptyset$ \EndIf \label{alg:evacuation:none}
    \State $U_{\min} \gets$ vertices in $U$ with the fewest moves from $v^i$ \label{alg:evacuation:fewest}
    \State $u_{\text{end}} \gets \arg\min_{u \in U_{\min}} h^i(u)$ \Comment{closest to the goal} \label{alg:evacuation:pick}
    \State \Return a shortest path $P$ from $v^i$ to $u_{\text{end}}$ over $E'$ \label{alg:evacuation:esc_return}
\EndProcedure
\Procedure{TakePath}{$i, P$} \Comment{$P = (u_0, \dots, u_m)$} \label{alg:evacuation:take_proc}
    \State $t_0 \gets \tau$; $t_k \gets t_{k-1} + \dur{i}{u_{k-1}}{u_k}$, $k = 1, \dots, m$ \label{alg:evacuation:times}
    \State $\partial a^i_{\text{next}} \gets (u_0, u_1, t_0, t_1)$ \Comment{first move now} \label{alg:evacuation:first}
    \State $\Phi^i \gets \langle (u_k, u_{k+1}, t_k, \bot) \rangle_{k=1}^{m-1}$ \Comment{action queue} \label{alg:evacuation:queue}
    \State \Return $t_m$ \label{alg:evacuation:return}
\EndProcedure
\end{algorithmic}
\end{algorithm}

\textbf{Evacuation.} \textsc{EscPath}$(i,\beta,L,d_{\max})$ (Alg.~\ref{alg:evacuation}) searches for an evacuation path
of agent $i$ with at most $d_{\max}$ moves. For strict evacuation, $L$ is the lane $L(\beta)$ and $d_{\max}$ is the parameter $d_s$. For loose evacuation, $L$
is empty and $d_{\max}$ is the parameter $d_l$.
 The search uses only the edges $E'$ whose swept areas do not collide with $\mathcal{C}^{-i}$, so agent $i$ pushes
nobody on the path (lines~\ref{alg:evacuation:edges}--\ref{alg:evacuation:bfs}). The set $U$ holds the reached
vertices whose footprints collide with neither $S(\beta)$ nor $L$ (line~\ref{alg:evacuation:goal}).
\textsc{EscPath} picks the vertex of $U$ with the fewest moves from $v^i$, and among these the one closest to the
goal of $i$ (lines~\ref{alg:evacuation:fewest}--\ref{alg:evacuation:pick}). It returns a shortest path $P$ to this
vertex $u_{\text{end}}$, or $\emptyset$ if $U$ is empty. \textsc{TakePath}$(i,P)$ lets the first move of $P$ start now and puts the
other moves into $\Phi^i$ with their start times
(lines~\ref{alg:evacuation:times}--\ref{alg:evacuation:queue}). It returns the arrival time at the end of $P$
(line~\ref{alg:evacuation:return}).

\section{Analysis}\label{lsrp-star-main:analysis}
LSRP*-LA keeps the guarantees of LSRP*~\cite{liu2026lsrpstar}.

\begin{theorem}[Completeness and optimality]\label{thm:guarantees}
LSRP*-LA terminates and finds a solution for every solvable instance. If it runs until $Open$ is empty, the
solution that it returns is optimal.
\end{theorem}
\begin{proof}[Proof sketch]
The theorem holds for LSRP*~\cite{liu2026lsrpstar}. LSRP*-LA differs from LSRP* only in how the action tree detects
conflicts and in the rule-based procedure. So it suffices to show that the following two properties still hold.

(i)~The leaves of $AT(a)$ are exactly the valid successors of $a$. A child of an action node $\theta$ adds a move only if the agent's swept area, shares no cell with $R_{\text{tree}}(\theta)$, so every leaf is valid.

(ii)~The rule-based procedure terminates and returns a valid successor. It terminates, because a pusher is planned
and only unplanned agents are pushed: no agent appears twice in a chain of pushes, so the recursion of
\textsc{AsyPush-LA} is at most $N$ deep. The successor is valid, because a move starts at the current event $\tau$
only if it collides neither with the cells occupied by the planned agents nor with the footprint of an unplanned
agent. So the assigned moves do not collide with each other. Moreover, no move enters the footprint of an agent that
is still unplanned, so this agent can always wait at its vertex, and every idle agent receives an action.
\end{proof}

Consider square agents on a
4-connected grid, and let $s^i$ denote the side length of agent $i$. Let agent $i$ move from $v^i$ to $v$ and push agent $j$ with ban chain $\beta=\langle(i,v)\rangle$, then $S(\beta)=F_{sw}^i(v^i,v)$. Let $i$'s lane $L(\beta)$ continue straight with length $\ell$. Then $T=S(\beta)\cup L(\beta)$, the cells
that $i$ covers in its next $\ell+1$ moves, is a straight strip of width $s^i$.

\begin{proposition}[Evacuation depth]\label{prop:depth}
(a) If no obstacle and no other agent is in its way, $j$ can leave $T$ by moving sideways within
$\lfloor (s^i+s^j)/2 \rfloor$ moves.
(b) To leave $T$ by moving ahead, $j$ needs at least $\ell+1$ moves.

\end{proposition}
\begin{proof}
(a) Let $j$ need $a$ sideways moves to leave $T$ on one side and $b$ on the other. A sideways move shifts $j$ by one
cell, so $a+b=s^i+s^j$ and $\min(a,b)\le\lfloor (s^i+s^j)/2\rfloor$, with equality if $|a-b|\le 1$, i.e., if $j$ is
centered across the width of $T$.
(b) $F^j(v^j)$ collides with $F^i(v)$ but not with $F^i(v^i)$, so $j$ stands directly ahead of $i$. The strip $T$
ends $\ell+1$ cells ahead of the front of $i$, so $j$ must move $\ell+1$ cells ahead.
\end{proof}

\noindent Recall that a strict and a loose evacuation have at most $d_s$ and $d_l$ moves, and that the lane covers
$\ell$ moves of the pusher. By (a), a pushed agent needs at most $\lfloor (s^i+s^j)/2\rfloor\le\max_i s^i$ sideways moves. So
$d_s,d_l\ge\max_i s^i$ makes an evacuation long enough to leave sideways. By (b), $\ell\ge d_s$ keeps a strict evacuation from moving ahead of the pusher. We use $\ell=6$, $d_s=4$ and
$d_l=3$, which meet both conditions for side lengths up to 3.

\section{Experimental Results}\label{lsrp-star-main:experiments}
\subsection{Experimental Setup}\label{sec:exp:setup}
We implement LSRP*-LA in C++ and run all experiments on a MacBook Air (Apple M4, 24\,GB RAM) with a 60\,s time limit
per instance. We use four maps from a MAPF benchmark~\cite{stern2019mapf}, den520d (open regions and narrow
passages), maze-128-128-10 (long corridors), Berlin\_1\_256 (a city map) and empty-48-48 (no obstacles), with 20
instances for each map and agent number $N$. Every agent is a square of $1{\times}1$, $2{\times}2$ or $3{\times}3$ cells with the cell of its vertex at the top left, and has one move duration from $\{1,2,\dots,5\}$; both are drawn uniformly. Starts and goals that do not fit are moved to a nearby free position, or the agent is shrunk, which leaves about 65\% of the agents larger than $1{\times}1$.

We compare against five baselines: (i) LSRP-LA, the rule-based procedure of LSRP*-LA without the search algorithm; (ii) CBS-AA-LA, CBS-AA~\cite{wu2026cbsaa} extended to large agents under our conflicts rule (Eq.~\eqref{eq:conflict}); (iii) PP-SIPP-LA, prioritized planning~\cite{silver2005cooperative} with
SIPP~\cite{phillips2011sipp} under our conflict rule; and the synchronous planners
(iv) MD-PIBT~\cite{jiang2026mdpibt} and (v) Layered LA-MAPF~\cite{yao2024layeredlamapf} with LA-LaCAM as its solver. As (iv) and (v) ignore the duration difference, we compare them separately in Sec.~\ref{sec:exp:async}.

\begin{figure}[tb]
  \centering
  \includegraphics[width=\columnwidth]{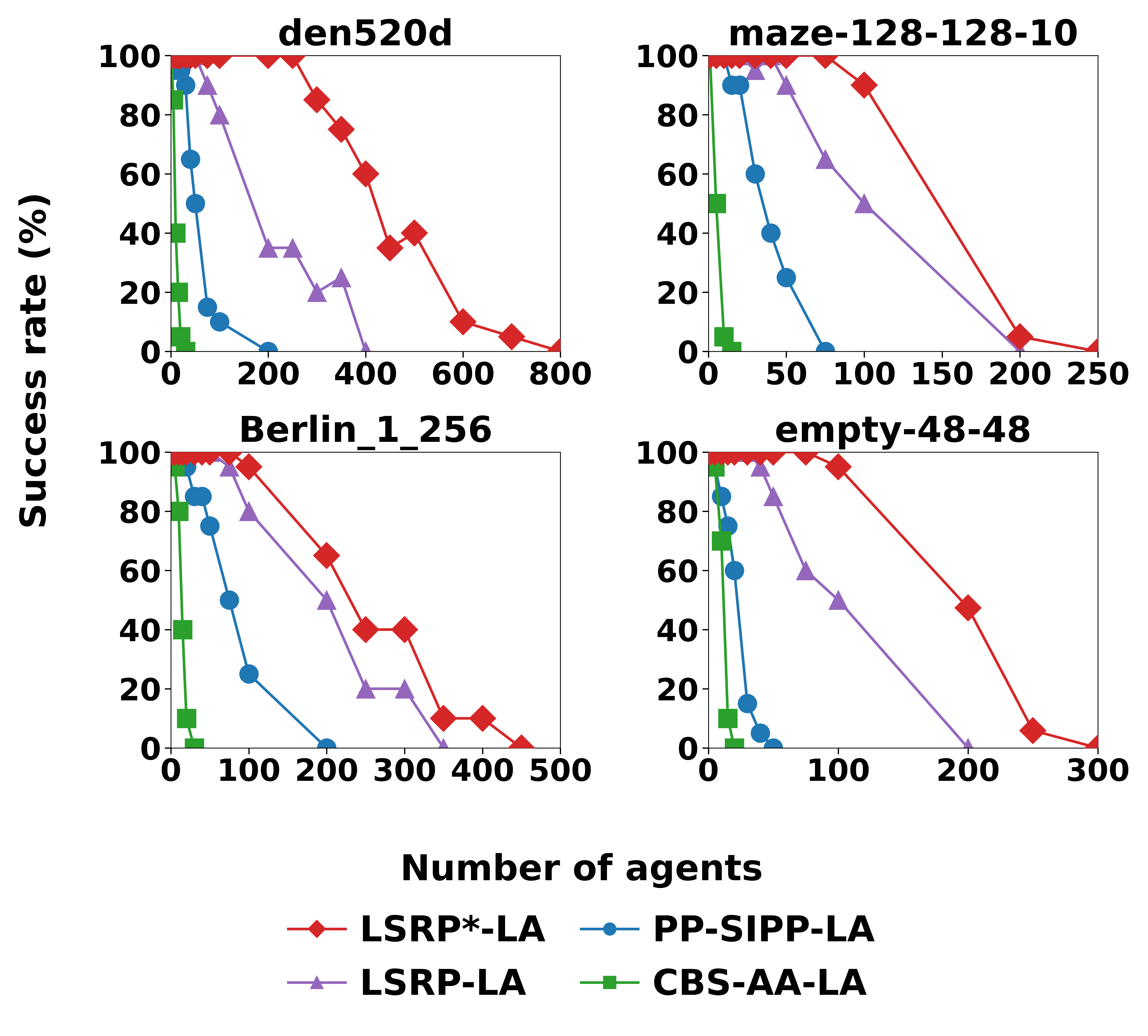}
  \caption{Success rates of the planners for LA-MAPF-AA (20 instances per $N$, 60\,s). LSRP*-LA scales the best on
all four maps.}
  \label{fig:success}
\end{figure}

\begin{figure}[tb]
  \centering
  \includegraphics[width=\columnwidth]{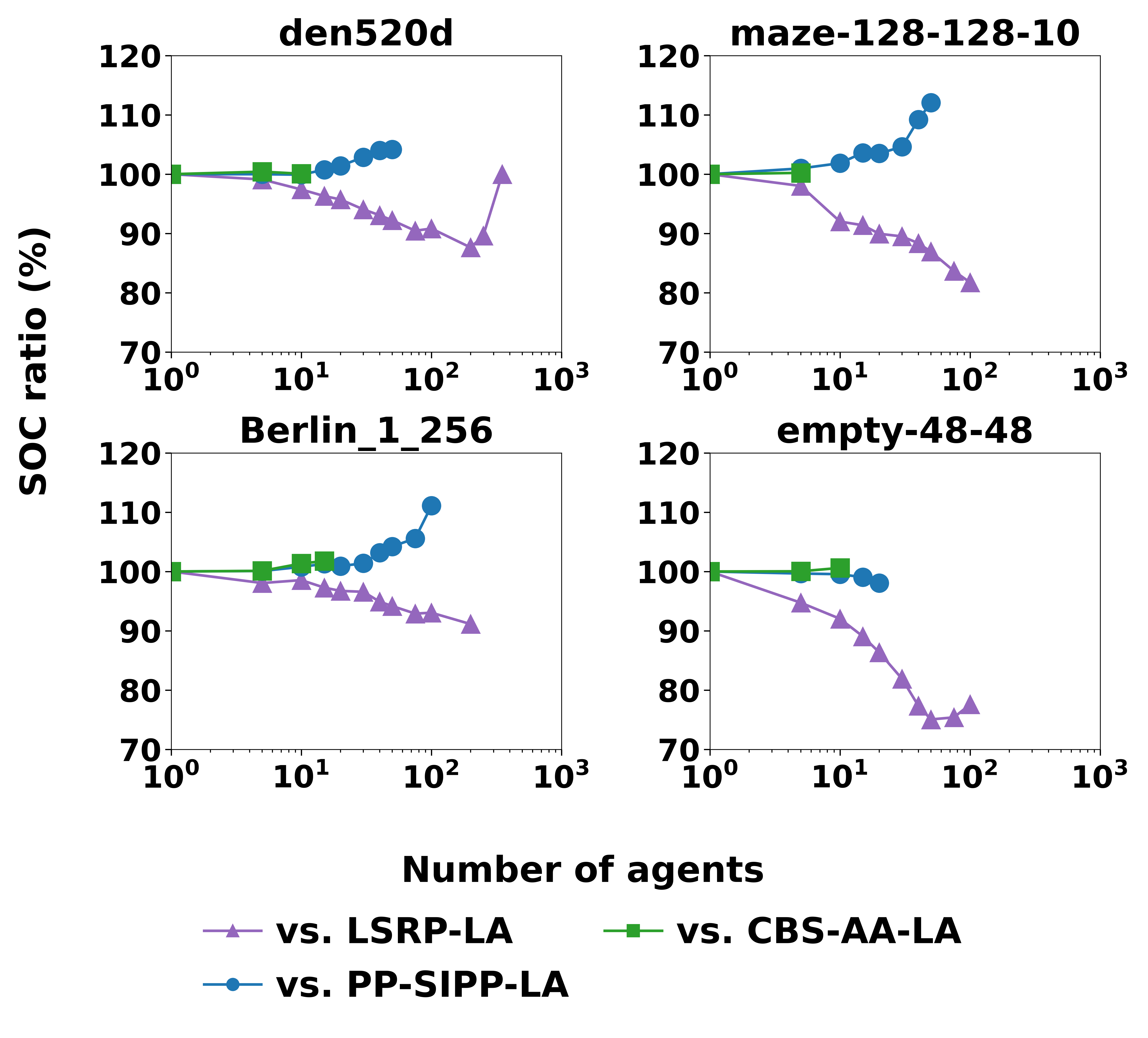}
 \caption{SOC ratio of LSRP*-LA over a baseline, computed on the instances solved by both. LSRP*-LA finds nearly optimal solutions (at most 2\% higher than CBS-AA-LA).}
  \label{fig:soc}
\end{figure}

\subsection{Success Rates}\label{sec:exp:success}
This experiment compares the planners for LA-MAPF-AA. We vary $N$ from 1 to 800 with a 60\,s time limit (Fig.~\ref{fig:success}). LSRP*-LA
scales best on all four maps. On den520d, it scales up to $N=800$, against $N=400$ for LSRP-LA, $N=200$ for
PP-SIPP-LA and $N=30$ for CBS-AA-LA, so the search doubles the number of agents that the rule-based procedure alone
can handle. On maze-128-128-10 and empty-48-48, LSRP*-LA scales up to $N=250$ and $N=300$: the maze has few alternative routes, and on the small empty map, 300 agents occupy more than half of the cells, causing the scalability lower compared to other maps.

\subsection{Solution Quality}\label{sec:exp:quality}
We compare the SOC of the final solutions with a 60\,s time limit. Each plot in Fig.~\ref{fig:soc} shows the SOC
ratio of LSRP*-LA over a baseline; lower is better. The purple curve (vs. LSRP-LA) drops below 100\% on all maps,
showing that LSRP*-LA finds solutions up to about 25\% cheaper than LSRP-LA. The green curve (vs. CBS-AA-LA) stays within 2\% of 100\%
on the small instances that CBS-AA-LA solves. As CBS-AA-LA extends an optimal planner, this shows that LSRP*-LA has high solution quality. The blue curve (vs. PP-SIPP-LA) is close to 100\% for small $N$ and
rises to about 112\% for larger $N$. It shows that LSRP*-LA obtains solution quality comparable to PP-SIPP-LA.

\begin{figure}[tb]
  \centering
  \includegraphics[width=\columnwidth]{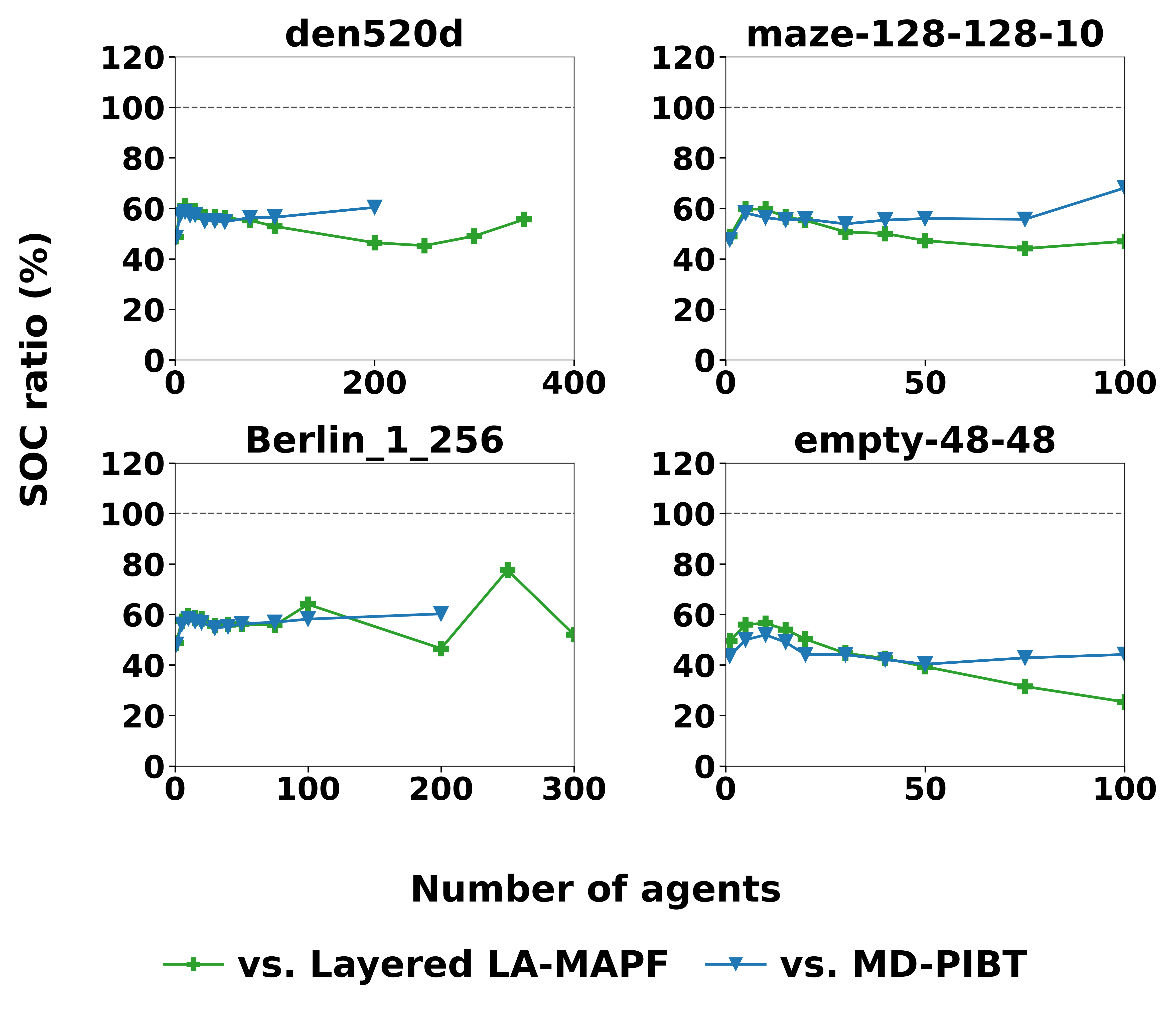}
  \caption{SOC ratio of LSRP*-LA over the synchronous planners executed with durations of 5. The dashed line marks equal SOC. LSRP*-LA yields solutions about half of synchronous planners.}
  \label{fig:x5}
\end{figure}

\subsection{Benefit of Asynchronous Planning}\label{sec:exp:async}
\begin{figure}[tb]
  \centering
  \includegraphics[width=\columnwidth]{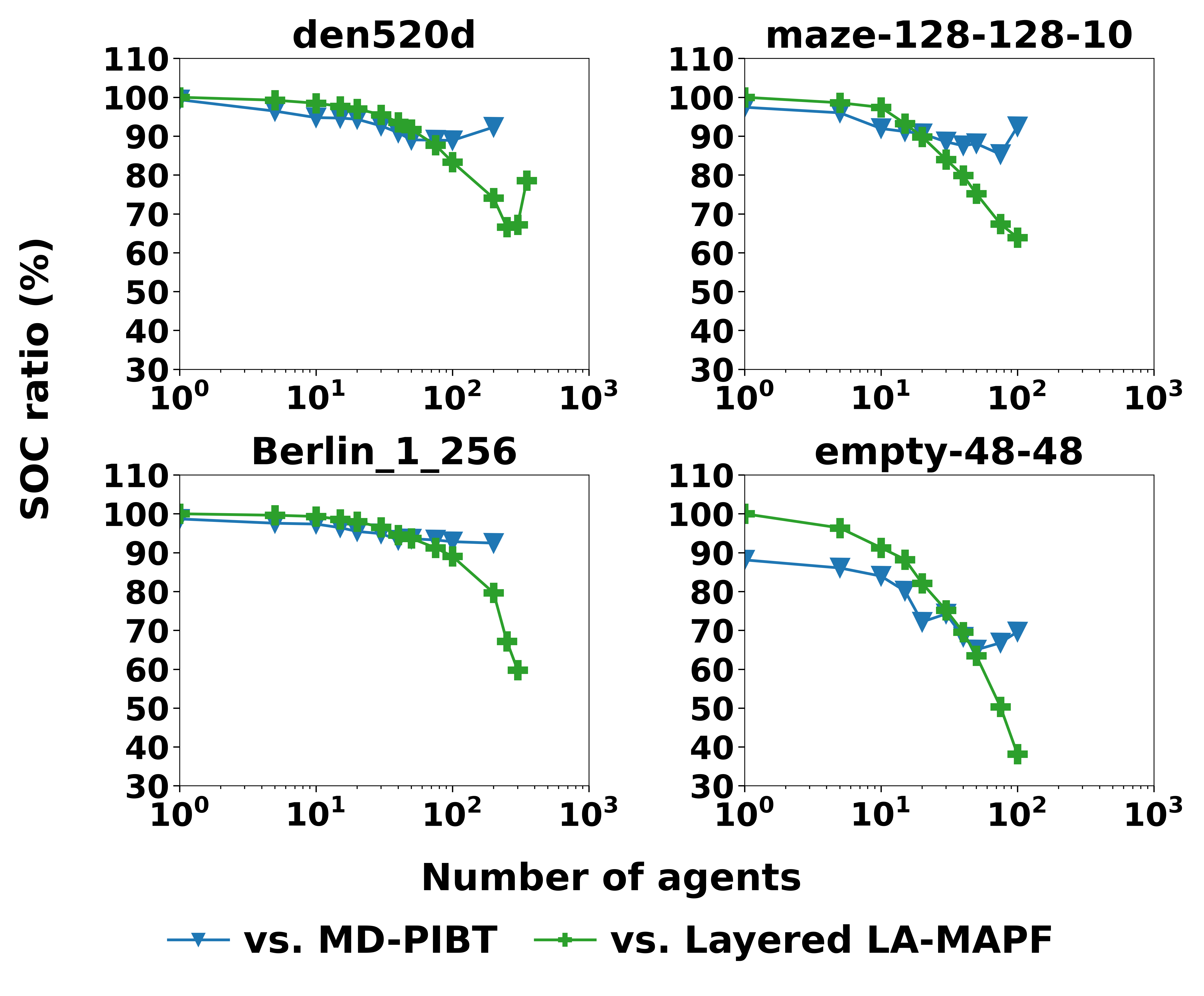}
  \caption{SOC ratio of LSRP*-LA over the synchronous planners in the synchronous setting. Even on the problem that these planners are designed for, LSRP*-LA is
never more expensive.}
  \label{fig:sync-soc}
\end{figure}

To solve LA-MAPF-AA, a synchronous planner must let every step last as long as the slowest action. Against MD-PIBT
and Layered LA-MAPF executed with durations of 5, LSRP*-LA costs about half of their SOC (25--78\%, Fig.~\ref{fig:x5}).
This comparison favors LSRP*-LA, since part of the gap comes from the durations of 5 alone: even for a single agent,
LSRP*-LA costs about half. To check whether the plans of LSRP*-LA are also better, we compare the planners on the
problem that the synchronous planners are designed for. We set all durations to 1, which turns the instances into
LA-MAPF instances (Remark~\ref{rem:special}), so the synchronous planners no longer lose time to the durations of 5.
LSRP*-LA is still never more expensive on average (Fig.~\ref{fig:sync-soc}): its SOC ratio is 65--99\% against
MD-PIBT and 38--100\% against Layered LA-MAPF. Hence the gain of LSRP*-LA does not come from asynchronous actions
alone; its plans are also cheaper.

\begin{figure}[tb]
  \centering
  \includegraphics[width=\columnwidth]{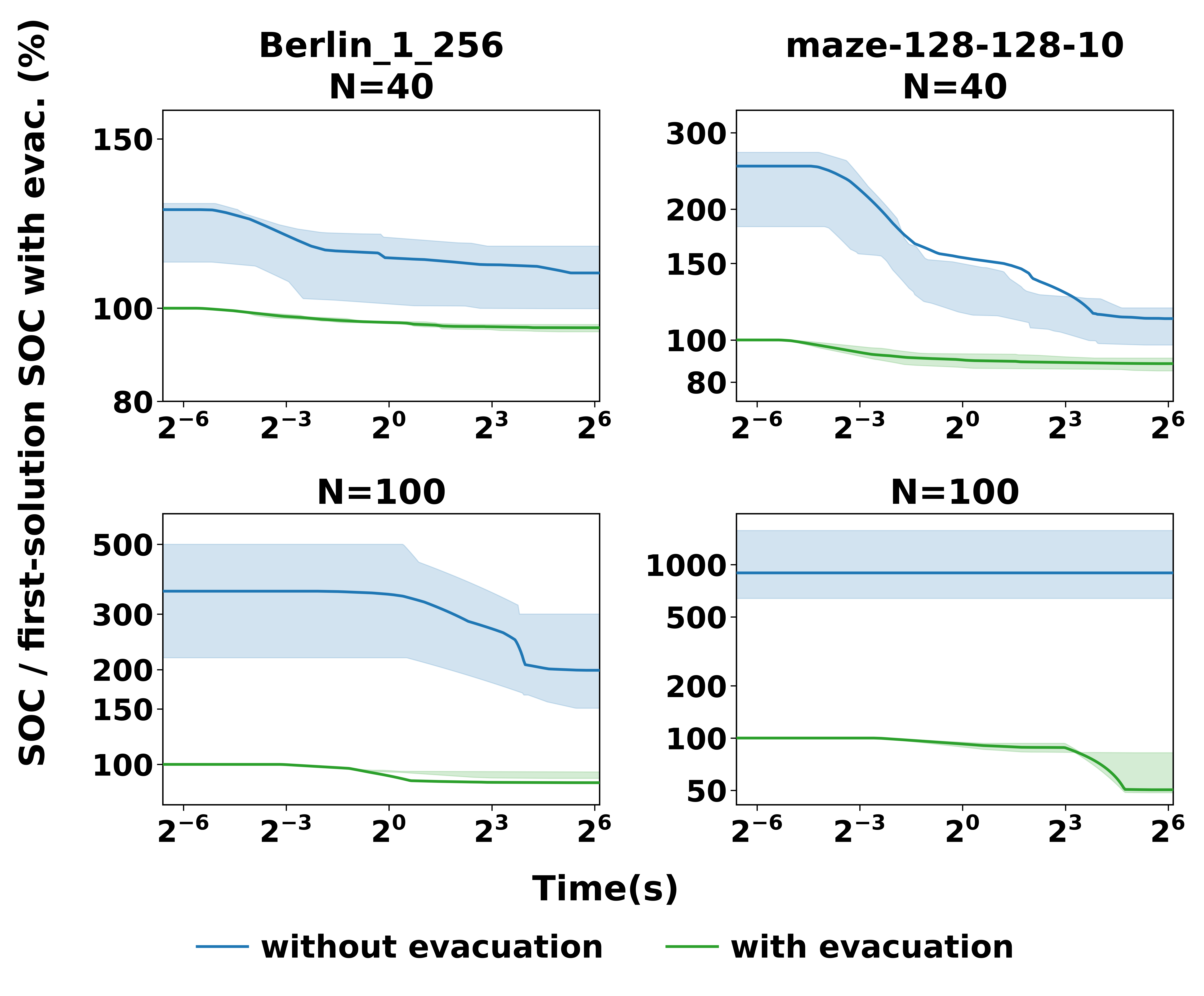}
  \caption{Effect of evacuation on anytime refinement. Shaded bands represent the 25--75\% range. Evacuation significantly improves the solution quality of LSRP*-LA.}
  \label{fig:evac}
\end{figure}

\subsection{Effect of Evacuation}\label{sec:exp:ablation}
We compare LSRP*-LA with and without evacuation on Berlin\_1\_256 and maze-128-128-10 with $N=40$ and $N=100$
(Fig.~\ref{fig:evac}). Without evacuation, a pushed agent makes a single move and is pushed again and again, so the
first solution costs 1.3 to 9.0 times as much. After 60\,s, the SOC without evacuation is still 110--900\% of the
first solution with evacuation, whereas with evacuation it falls to 50--95\%.

\section{Conclusion and Future Work}\label{lsrp-star-main:conclusion}
This paper extends LSRP* to LSRP*-LA, a scalable, complete and anytime planner for MAPF with large agents and
asynchronous actions. Its evacuation lets pushed agents make way for large agents, and it scales to hundreds of
agents with high-quality solutions. Future work includes agents that rotate when pushed, to move aside in narrow spaces.

\bibliographystyle{IEEEtran}
\bibliography{references}

\end{document}